\documentclass[letterpaper, 10 pt, conference]{ieeeconf} 
\pdfoutput=1
\IEEEoverridecommandlockouts

\usepackage[utf8]{inputenc} 

\usepackage{amsmath, amssymb, amscd}
\usepackage{wasysym} 
\usepackage{amsfonts}
\usepackage{nicefrac}
\usepackage{bm}

\usepackage[footnotesize]{caption}
\usepackage{graphicx}

\let\labelindent\relax 
\usepackage{enumitem}
\usepackage{multirow}
\makeatletter
\let\NAT@parse\undefined
\makeatother
\usepackage[numbers,sort&compress]{natbib}
\usepackage{multicol}
\usepackage[bookmarks=true]{hyperref}
\usepackage{booktabs}

\newcommand\blfootnote[1]{%
  \begingroup
  \renewcommand\thefootnote{}\footnote{#1}%
  \addtocounter{footnote}{-1}%
  \endgroup
}

\title{\LARGE \bf Differentiable Factor Graph Optimization for Learning Smoothers}

\author{Brent Yi$^{1}$, Michelle A. Lee$^{1}$, Alina Kloss$^{2}$, Roberto Mart\'in-Mart\'in$^{1}$, and Jeannette Bohg$^{1}$}

\begin{document}

\twocolumn[{%
    \renewcommand\twocolumn[1][]{#1}%
    \maketitle
    \vspace{-10pt}
    \newcommand{\teaserwidth}{\textwidth}
    \includegraphics[width=\teaserwidth]{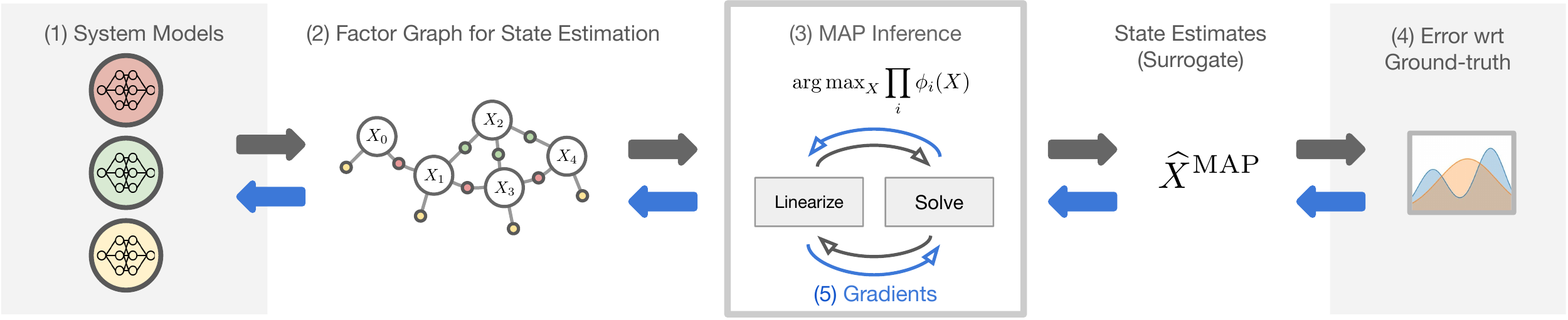}
    \captionof{figure}{
        \label{fig:teaser}
        \textbf{An overview of our approach for end-to-end learning of factor graph-based smoothers.}
        We \textbf{(1)} start with probabilistic system models, which may be learned or analytical, \textbf{(2)} use them to model state estimation problems as factor graphs, \textbf{(3)} generate a surrogate for posterior estimates by running several steps of a nonlinear optimizer, \textbf{(4)} compute an error with respect to a known ground-truth, and \textbf{(5)} backpropagate estimation errors through the unrolled nonlinear optimizer directly to system model parameters.
    }%
}]

\begin{abstract}
A recent line of work has shown that end-to-end optimization of Bayesian filters can be used to learn state estimators for systems whose underlying models are difficult to hand-design or tune, while retaining the core advantages of probabilistic state estimation.
As an alternative approach for state estimation in these settings, we present an end-to-end approach for learning state estimators modeled as factor graph-based smoothers.
By unrolling the optimizer we use for maximum a posteriori inference in these probabilistic graphical models, we can learn probabilistic system models in the full context of an overall state estimator, while also taking advantage of the distinct accuracy and runtime advantages that smoothers offer over recursive filters.
We study this approach using two fundamental state estimation problems, object tracking and visual odometry, where we demonstrate a significant improvement over existing baselines.
Our work comes with an extensive code release, which includes training and evaluation scripts, as well as Python libraries for Lie theory and factor graph optimization: \url{https://sites.google.com/view/diffsmoothing/}.
\end{abstract}

\blfootnote{$^1$ Stanford University, \{brentyi, michellelee, robertom, bohg\}@cs.stanford.edu}
\blfootnote{$^2$ Max Planck Institute for Intelligent Systems, akloss@tue.mpg.de}


\newcommand{\factorpotential}{\bm{\phi}}
\newcommand{\factorresidual}{\mathbf{r}}

\newcommand{\processcov}{\mathbf{Q}}
\newcommand{\measurementcov}{\mathbf{R}}
\newcommand{\dynamics}{\mathbf{f}}
\newcommand{\virtualsensor}{\mathbf{g}}
\newcommand{\rawimage}{\mathbf{I}}
\newcommand{\measurement}{\mathbf{z}}

\newcommand{\ekfmu}{\hat{\mathbf{x}}}
\newcommand{\ekfsigma}{\mathbf{P}}
\newcommand{\dynamicsjacobian}{\mathbf{A}}
\newcommand{\kalmangain}{\mathbf{K}}
\newcommand{\measurementjacobian}{\mathbf{H}}
\newcommand{\measurementmodel}{\mathbf{h}}
\newcommand{\identity}{\mathbb{I}}
\section{Introduction}



%
%

State estimation is a universal problem in robotics. To make informed decisions, nearly all robots, regardless of where they are deployed, must be able to infer system state from noisy sensor observations. 
This problem is often addressed using methods such as
 Kalman or particle filters~\citep{thrun2005probrob}, as well as optimization-based smoothers~\citep{FactorGraphBook}.
These algorithms are rooted in probability theory, and require uncertainty-aware models to describe state transitions and how sensor observations relate to system state.
The probabilistic underpinnings of these methods have a wide range of practical advantages: information about uncertainty can be used not only for human interpretability of the inner workings of the state estimator, but also as input to risk-aware planning or control algorithms~\cite{RiskSensitive:Todorov:2007,Ponton2020,eppner2018physics}.

In many applications, models for these estimators are easy to derive analytically.
For pose estimation with inertial measurement units, physically-grounded sensor models can relate angular velocities from a gyroscope, linear accelerations from an accelerometer, and absolute orientations from a magnetometer to information about an egocentric pose.
For an inverted pendulum, precise equations of motion can be derived using Newtonian or Lagrangian mechanics.
Uncertainty in systems like these can often be effectively approximated as Gaussians with manually-tuned, diagonal covariance matrices.

More complex robotic systems, however, often contain sensors or dynamics that are much harder to pre-specify or tune.
This is particularly true for some of the richest, most information-dense sensing modalities, such as images, audio, or tactile feedback, as well as in interaction-rich applications that need to reason about difficult-to-model contact dynamics.
These systems also tend to have more complex, heteroscedastic (variable) noise profiles~\cite{kloss2018howtotrain}.
For example, an object position estimate from an image might rapidly switch from being extremely precise under nominal operating conditions to completely useless under occlusion or poor lighting.


In this work, we address end-to-end learning for probabilistic state estimators modeled as factor graph-based smoothers.
Learned smoothers offer the option of circumventing analytical models without sacrificing the advantages of probabilistic state estimation.
Smoothers also have distinct accuracy and runtime benefits over recursive filters, particularly for more complex or nonlinear applications.
Improved integration of smoothing into learning-based systems promises to open these benefits up to a broader set of estimation problems.

The contributions of our work are:
\begin{itemize}
    \item An approach for end-to-end learning of smoothing-based state estimators, using a surrogate loss computed via a sequence of differentiable optimization steps.
    \item A study on learning sensor and dynamics models with both constant and heteroscedastic uncertainties, using standard experiments from past differentiable filtering works: 2D visual object tracking and visual odometry. 
    \item An empirical comparison of learnable state estimators and loss functions, where the proposed approach outperforms alternatives.
    \item The release of open-source Python libraries targeted at differentiable optimization, written using JAX~\cite{jax2018github}.
\end{itemize}

%

\section{Background \& Related Work}
\label{sec:related_work}

\subsection{State Estimation}
In robotics, state estimation has primarily been studied under two frameworks.

\subsubsection{Filtering}
State estimators are commonly built using a filtering structure~\cite{thrun2005probrob}.
In a filter, we maintain a probabilistic belief about the current state of the system, which is updated at each timestep using stochastic dynamics and observation models.
This process requires approximations for nonlinear systems, but is recursive in nature and only requires a constant amount of memory for problems with a fixed state dimension: the belief at a given timestep $t$ can be computed from only the belief at $t - 1$ and any observations or control inputs at timestep $t$.

\subsubsection{Smoothing}
In contrast to filtering, smoothing retains all past measurements and continuously recovers full state trajectories.
While some formulations of smoothers (such as Kalman smoothers) rely on similar approximations and belief representations as their filtering-based counterparts, prior works have shown that by circumventing the need for explicitly parameterizing belief distributions, smoothers based on iterative optimization can outperform filtering approaches in both accuracy and runtime~\cite{dellaert2006squareroot,strasdat2012visual}. 
We elaborate on these benefits in Appendix \ref{appendix:whysmoothing}.

In this work, we build on top of a common interface for optimization-based smoothers: factor graphs. 
In state estimation, these probabilistic graphical models have seen the most adoption in SLAM applications~\cite{salasmoreno2013,leutenegger2015,whelan2015,isam,isam2}, but have more recently also found use cases in areas like tactile sensing~\cite{yu2018,lambert2019visuotactile,GelSightFAIR} and multi-object tracking~\cite{Pschmann2020FactorGB}.

\subsection{Differentiable Filtering}

Probabilistic state estimators traditionally assume that analytical models for observations and transitions are available and tractable to evaluate.
In order to better integrate learned models into filters, several works have explored backpropagation through the prediction and correction steps of a Bayesian filter.
The parameters of the system models, which may otherwise be difficult to hand-design or tune, can then be learned to directly minimize end-to-end estimation error.
For visual tracking and robot localization tasks, \citet{haarnoja2016backprop} propose a Kalman filtering-based approach, while \citet{karkus2018particle} and  \citet{jonschkowski2018differentiable} evaluate differentiable flavors of the particle filter. 
Meanwhile, \citet{kloss2018howtotrain}, \citet{lee2020}, and \citet{zachares2021interpreting} have each applied differentiable filters to applications in robot manipulation. 

\subsection{Combining Learning \& Smoothers}
  
Various works have also explored adding learned models to factor graphs and smoothers.
In contrast to our work, which is focused on end-to-end learning, \citet{czarnowski2020deepfactors}, \citet{IV-SLAM}, and \citet{GelSightFAIR} each learn parameters outside of the smoothing formulation, which are then evaluated in an optimization-based smoother.
\cite{czarnowski2020deepfactors} uses depth maps predicted by a neural network to compute photometric, reprojection, and sparse geometric factors in a monocular SLAM system.
\cite{IV-SLAM} proposes learning a context-dependent noise model to guide feature extraction for SLAM.
\cite{GelSightFAIR} incorporates tactile measurements into a factor graph to estimate manipulated object poses.

Prior works have explored backpropagation through a Kalman-Bucy smoother for learning continuous-time stochastic dynamics \cite{liwu2021replayovershooting} and applied proximal gradient optimization techniques for minimizing prediction errors on Kalman smoothers \cite{barratt2020fitting}, but end-to-end learning has not been studied for smoothers based on iterative optimization.

\subsection{Differentiable Optimization}
\label{sec:diff_opt}

Various algorithms have recently been developed for treating optimization itself as a differentiable operation.
\citet{tang2018banet} and \citet{gradslam2020} propose differentiable versions of the Levenberg-Marquardt solver, with the former taking a prediction approach to computing the damping coefficient and the latter choosing a soft reparameterization of the damping coefficient.
\citet{gn-net-2020} formulate a probabilistic loss based on a single-step Gauss-Newton update.
For differentiable optimization through first-order methods, \citet{grefenstette2019generalized} presents tools for unrolling differentiable optimizers in PyTorch.

In the context of factor graphs, \citet{bhardwaj2020dgpmp} backpropagate through Gauss-Newton updates to learn parameters for Gaussian Process Motion Planning, which formulates trajectory optimization as inference on a factor graph.
Our work builds off similar ideas, but is targeted at state estimation. 

Proir works have largely relied on dense matrix computations with manually specified Jacobians.
In contrast, our code release includes differentiable optimizers backed by sparse matrix computations, with automatically differentiated Jacobians.

\section{Method}
\label{sec:approach}



In this work, we present \textit{differentiable factor graph optimization} for learning smoothers. 
By interpreting the nonlinear optimization steps used for maximum a posteriori (MAP) inference on a factor graph as a sequence of differentiable operations, our approach brings end-to-end learning to probabilistic smoothers.


\subsection{State Estimation with Factor Graphs}
\label{sec:factor_graphs}

Factor graphs are a class of probabilistic graphical models that have demonstrated impressive versatility in a wealth of real-world robotics problems.
These bipartite graphs leverage conditional independence structures to represent probability density over sets of \textit{variables} constrained by \textit{factors}, where factors define conditionally independent density functions on subsets of the variables.

For state estimation problems, we represent the quantities $X$ we want to estimate as variable nodes, and information from system models and sensor observations as factor nodes.
Assuming Gaussian noise, each factor encodes a joint probability density for a specific value of $X$ as follows:
\begin{equation}
    \label{eq:factor_potential_definition}
    \factorpotential^\theta_i (X) \propto \exp(-\frac{1}{2} \lvert\lvert \mathbf{r}^\theta_i(X) \rvert\rvert_{\Sigma^\theta_i}^2)
\end{equation}
where $\lvert\lvert \cdot \lvert\lvert^2_{\Sigma_i^\theta}$ denotes the squared Mahalanobis norm of an error vector $\factorresidual^\theta_i(X)$ under covariance $\Sigma_i^\theta$.
$\theta$ contains all factor parameters that can be learned or tuned, which may include covariance matrices, neural network weights, or physical quantities like masses and inertia matrices.
In this paper, all terms (values or functions) conditioned on these parameters will be superscripted with $\theta$.


The densities specified by each factor can be used to solve for an estimate of the unknown states $X$ -- the set of all variables in the factor graph -- by performing maximum a posteriori (MAP) inference over the factor graph.
Assuming Gaussian noise, this problem reduces to a nonlinear least squares optimization:
\begin{equation}
    \label{eq:map_definition}
    \begin{split}
        X^{\text{MAP}, \theta}
        &= {\arg\max}_X \prod_i \factorpotential_i^\theta(X)\\
        &= {\arg\max}_X \prod_i \exp(-\frac{1}{2} \lvert\lvert
            \factorresidual_i^\theta(X)
        \rvert\rvert_{\Sigma_i^\theta}^2)\\
        &= {\arg\min}_X \sum_i \lvert\lvert
            {\Sigma_i^\theta}^{-\frac{1}{2}} \factorresidual_i^\theta(X)
        \rvert\rvert_2^2\\
    \end{split}
\end{equation}

These nonlinear optimization problems are typically addressed with approaches like the Gauss-Newton or Levenberg-Marquardt algorithms, which solve a sequence of linear subproblems around first-order Taylor approximations of the cost function.




\subsection{End-to-end Factor Learning}
%
The key insight that our work builds on is that multi-component systems whose parameters have been optimized end-to-end can perform better than systems whose components are optimized in isolation and brought together~\citep{KarkusMHKLL19}.
As a substitute for individually tuning each factor in a factor graph, we study end-to-end optimization of the mean-squared error between ground-truth labels $X^\text{gt}$ and the factor graph's MAP estimate $X^{\text{MAP}, \theta}$:

\begin{equation}
    \label{eq:mse_loss}
    \mathcal{L}_{\text{mse}}(\theta)
    = \sum_t \lvert\lvert \text{diag}(\bm{\alpha}) (X_t^{\text{gt}} \ominus X^{\text{MAP},\theta}_t) \rvert\vert_2^2
\end{equation}
where $t$ is an index corresponding to each variable in the factor graph and $\bm{\alpha}$ is a hyperparameter vector that can be used to weight components of the error vectors.
$\ominus$ is a generalized subtraction operator, which enables losses for variables on non-Euclidean manifolds like Lie groups. 

To enable backpropagation of this loss directly to learnable parameters within the factors, an initial step is to reconstruct MAP inference as a fully differentiable computation graph.
Solving for $X^{\text{MAP}, \theta}$, however, typically requires running a nonlinear optimizer until a convergence criteria is met.
Not only are these problems generally not convex, but the sheer number of iterations that convergence might require means that naively implementing a differentiable nonlinear optimizer and backpropagating the loss in Equation \ref{eq:mse_loss} would be computationally prohibitive.
For this reason, we propose a surrogate loss $\widehat{\mathcal{L}}_\text{mse}(\theta)$ that is identical to Equation \ref{eq:mse_loss} but replaces supervision on the MAP estimate $X^{\text{MAP}, \theta}$ with supervision on a surrogate $\widehat{X}^{\text{MAP}, \theta}$.

We motivate the computation of $\widehat{X}^{\text{MAP}, \theta}$ with two requirements.
First, the surrogate should converge to $X^\text{gt}$ when the factor parameters have been correctly learned.
Second, it must be practical for integration into a mini-batch stochastic gradient descent pipeline.

To fulfill these criteria, we propose computing $\widehat{X}^{\text{MAP}, \theta}$ by (a) initializing a nonlinear optimizer with the ground-truth trajectory and (b) unrolling a constant number of $K$ nonlinear optimization steps \textit{away} from the ground-truth label, where the last step's output is supervised in the surrogate loss.
During each nonlinear optimization step in a forward pass through steps $k = 0 \dots K - 1$, the computation graph linearizes the MAP cost (Equation \ref{eq:map_definition}) around the current variable values --- this outputs a sparse Jacobian --- and computes a local update $\Delta^\theta_k$ by solving the resulting sparse linear subproblem.
As an example for $K = 3$, we get:
\begin{equation}
    \label{eq:surrogate_loss_K_eq_3}
    \widehat{X}^{\text{MAP}, \theta}
    = ((X^\text{gt} \oplus \Delta^\theta_0) \oplus \Delta^\theta_1) \oplus \Delta^\theta_2
\end{equation}
where $\oplus$ is a generalized addition operator. 

After applying this sequence of standard nonlinear optimization steps (for example, Gauss-Newton), the surrogate loss can be directly minimized:
\begin{equation}
    \widehat{\mathcal{L}}_{\text{mse}}(\theta) = \sum_t \lvert\lvert \text{diag}(\bm{\alpha}) (X_t^{\text{gt}} \ominus \widehat{X}^{\text{MAP},\theta}_t) \rvert\vert_2^2
\end{equation}

This formulation is desirable for several practical reasons.
Initializing our optimizer with the ground-truth trajectory sidesteps concerns about convergence: it guarantees that when the target model parameters perfectly match the ground-truth data, $X^{\text{MAP}, \theta} = X^\text{gt}$, all local update values $\Delta^\theta_k$, computed errors, and backpropagated gradients evaluate to zero.
Unrolling a fixed number of iterations also has significant advantages for GPU-bound operations.
It simplifies batching, eliminates branch and loop overheads, and has a static memory footprint that enables reverse-mode auto-differentiation using tools like Tensorflow XLA \cite{jax2018github}.

While we frame this differentiable optimization process as inference on factor graphs, note that the underlying computations are not tied to this particular probabilistic graphical model. 
This same approach can be applied to the broad range of similar optimization problems that appear when visualizing these systems as hypergraphs \cite{kummerle2011g2o}, where factor nodes are replaced by hyperedges that can connect any number of variable nodes, and in application-specific algorithms for problems like pose graph optimization \cite{OlsonGraph2006} and bundle adjustment \cite{strasdat2012visual}.

\section{Experimental Setup}
\label{sec:experimental_setup}

\subsection{Visual Tracking}\label{sec:vis_track}

\newcommand{\diskpos}{\mathbf{p}}
\newcommand{\diskvel}{\mathbf{v}}
\newcommand{\diskq}{\mathbf{q}}

Our first set of experiments study a synthetic visual tracking environment that has been used to evaluate differentiable filters \cite{haarnoja2016backprop, kloss2018howtotrain}.
The goal of this task is to track the motion of a red disk as it follows linear-Gaussian dynamics in 2D space, given images inputs that contain occasional occlusions by other discs.
An example observation can be viewed in Figure~\ref{fig:disk_visualization}, and details can be found in Appendix \ref{appendix:experiment_visual_tracking}.

The factors used for this task are formalized below.

\begin{figure}[t!]
    \centering
    \vspace{0.1em}
    \includegraphics[width=\linewidth]{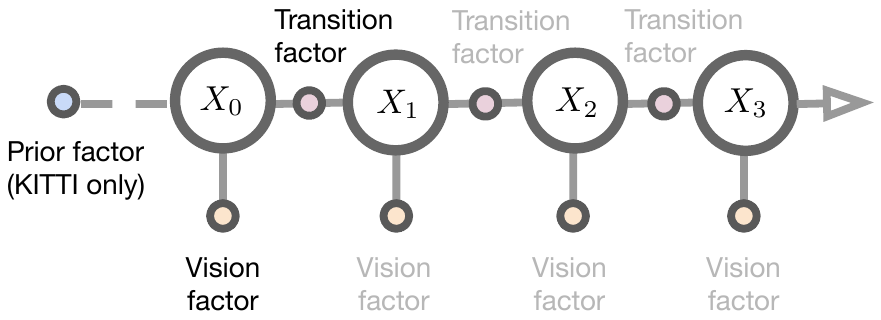}
    \caption{
        \textbf{Factor graph representation of the learned smoothers.}
        In addition to state variable nodes $X_0 \dots X_{T - 1}$, graphs also contain \textit{transition factors} that encode system dynamics, \textit{vision factors} for processing raw images, and for the KITTI visual odometry task \textit{prior factors} that constrain start states.
    }
    \label{fig:visual_tracking_factor_graph}
\end{figure}

\subsubsection{Transition Factors}
These factors use the known disk dynamics model (see Appendix \ref{appendix:experiment_visual_tracking}) to define joint densities over variable pairs from temporally adjacent timesteps.
Given the process noise covariance $\processcov_\text{disk}$, forward dynamics model $\dynamics$, and notation from Equation \ref{eq:factor_potential_definition}, we define the transition factor as:
\begin{equation}
        \factorresidual_i(X_t, X_{t + 1}) = \dynamics(X_t) - X_{t+1},\quad
        \Sigma_i = \processcov_\text{disk}
\end{equation}

\subsubsection{Learned Vision Factors}
Hand-designing models for processing high-dimensional image inputs is challenging. Therefore, we use a factor with learned parameters for this task.
The core component of this factor is a probabilistic virtual sensor model $\virtualsensor^\theta$, which is shared across all vision factor instantiations.
This virtual sensor maps a raw image observation $\rawimage_t$ to a position measurement $\measurement_t^\theta$ and measurement covariance $\measurementcov_t^\theta$:
\begin{equation}
    (\measurement^\theta_t, \measurementcov^\theta_t) = \virtualsensor^\theta(\rawimage_t)
\end{equation}
We can define the resulting factor as:
\begin{equation}\label{eq:virtual_sensor_disk}
    \factorresidual^\theta_i(X_t=(\diskpos_t, \diskvel_t)) = \measurement^\theta_t - \diskpos_t
    ,\quad
    \Sigma^\theta_i = \measurementcov^\theta_t
\end{equation}

We use the proposed method to learn virtual sensor models with constant $\measurementcov_t^\theta$ noise profiles, as well as models that output image-dependent heteroscedastic noise.
The network architecture used for the heteroscedastic noise model can be found in Appendix \ref{appendix:network_architectures}.

%

%

\subsection{Visual Odometry}\label{sec:vis_odo}

\newcommand{\kittipose}{\mathbf{T}} 
\newcommand{\kittilinearvel}{\mathbf{v}}
\newcommand{\kittiangularvel}{\bm{\omega}}

Again mirroring prior filtering works, we apply our approach to the \texttt{kitti-10} visual odometry task described in \cite{kloss2018howtotrain}.
This task uses real-world data with nonlinear dynamics.

The goal of this task is to use image observations for estimating the 5 degree-of-freedom state of a vehicle moving through the world.
We represent the variable nodes as $X_t = (\kittipose_t, \kittilinearvel_t, \kittiangularvel_t)$ where
 $\kittipose \in \mathrm{SE}(2)$ is a 2D transformation between the world and vehicle body frames, $\kittilinearvel$ is a scalar forward velocity, and $\kittiangularvel$ is a scalar angular velocity. 
This is challenging: not only must the state estimator learn to operate using only noisy visual observations, but it also needs to do so in a way that generalizes to novel environments.
The validation trajectories are taken exclusively from never-before-seen roads.


The factors used for this task are formalized below.

%
%

\subsubsection{Transition Factors}
As with many robot odometry tasks, an analytical transition model $\dynamics$ can be accurately hand-specified for the KITTI task. 
The transition factors take neighboring variable pairs as input and defines two costs: (1) differences between actual and predicted poses at $t + 1$ and (2) changes in velocity.
By defining the dynamics as a velocity integrator, $\mathrm{SE}(2) \times \mathbb{R}^2 \stackrel{\dynamics}{\to} \mathrm{SE}(2)$, the transition factor can be written as:
\begin{equation}\label{eq:kitti_dyn}
    \begin{split}
        \factorresidual_i(
            X_t, X_{t+1}
        )
        &= \begin{bmatrix}
            \log(\dynamics(\kittipose_t, \kittilinearvel_t, \kittiangularvel_t)^{-1} \kittipose_{t+1})^\vee\\
            \kittilinearvel_{t+1} - \kittilinearvel_t\\
            \kittiangularvel_{t+1} - \kittiangularvel_t
        \end{bmatrix}
    \end{split}
\end{equation}
where $\mathrm{SE}(2) \stackrel{\log}{\to} se(2)$ and $se(2) \stackrel{(\cdot)^\vee}{\to} \mathbb{R}^3$.
We defer to \cite{sola2018microlie} for a comprehensive introduction to Lie theory for robotics.

The dynamics and residual computations have no learnable parameters, but all transition factors share a learnable noise covariance $\Sigma_i^\theta = \processcov^\theta_\text{KITTI}$.
We assume the driver's actions to be unknown and they are therefore included in this noise model.


\subsubsection{Learned Vision Factors}
Because a vision system on a vehicle must generalize to never-before-seen roads --- as is the case in our withheld validation trajectory --- it is not possible to determine an absolute vehicle pose from visual data.
Instead, we follow the formulation established in prior works \cite{haarnoja2016backprop,karkus2018particle,jonschkowski2018differentiable,kloss2018howtotrain} and augment observations for predicting velocities.
We stack each raw RGB image channel-wise with a difference frame between the current and previous timesteps.
Using this input, we learn a virtual sensor model $\virtualsensor^\theta$ that outputs instantaneous linear and angular velocities:
\begin{equation}
    (\measurement^\theta_{\kittilinearvel, t},\ \measurement^\theta_{\kittiangularvel, t},\ \measurementcov^\theta_t) = \virtualsensor^\theta(\rawimage_t)
\end{equation}

Then, the resulting factor can be defined as:
\begin{equation} \label{eq:virtual_sensor_odo}
    \begin{split}
        \factorresidual^\theta_i(X_t=(\kittipose_t, \kittilinearvel_t, \kittiangularvel_t))
        &= \begin{bmatrix}
            \measurement^\theta_{\kittilinearvel, t} - \kittilinearvel_t
            & \measurement^\theta_{\kittiangularvel, t} - \kittiangularvel_t
        \end{bmatrix}^\top\\
        \Sigma^\theta_i
        &= \measurementcov^\theta_t
    \end{split}
\end{equation}

We again run experiments using virtual sensors with both constant and heteroscedastic noise profiles.

\subsubsection{Prior factors}
Because our dynamics and vision factors for this task are only useful for determining relative poses, we additionally introduce an absolute \textit{prior factor}.
This factor constrains the estimated trajectory by anchoring the position and velocity of the vehicle at the very first timestep:
\begin{equation}
    \begin{split}
        \factorresidual_i(X_0)
        &= \begin{bmatrix}
            \log((\kittipose_0^\text{gt})^{-1} \kittipose_0)^\vee\\
            \kittilinearvel^\text{gt}_0 - \kittilinearvel_0\\
            \kittiangularvel^\text{gt}_0 - \kittiangularvel_0\\
        \end{bmatrix}
    \end{split}
\end{equation}
A diagonal covariance is used.
Because this is the only factor that constrains the estimated trajectory's absolute location, the position terms can be defined by any (numerically stable) positive definite matrix with no impact on converged MAP estimates.
To emulate a hard constraint, the velocity terms are parameterized using a square-root precision value of $1e7$.

\subsection{Baselines}

We present results for three sets of baselines: an alternative probabilistic approach for learning factor parameters, differentiable filters, and LSTM approaches for state estimation.

\subsubsection{Joint NLL Loss}
For learning parameters of the factors, one natural loss function would be to find the parameters that minimize the negative log-likelihood of each ground-truth trajectory $X^\text{label}$ over the joint distribution specified by the factor graph:
\begin{equation} \label{eq:nll}
    \begin{split}
        \mathcal{L}_{\text{nll}}(\theta)
        &= -\log \prod_i \factorpotential^\theta_i(X^{\text{gt}})\\
        &= -\sum_i \log (\factorpotential^\theta_i(X^{\text{gt}}))\\
        &= -\sum_i \left(
            \log \lvert\lvert (\factorresidual^\theta_i(X^\text{gt})) \rvert\rvert^2_{\Sigma_i^\theta}
            + \log \lvert \Sigma_i^\theta \rvert
        \right)\\
    \end{split}
\end{equation}
where constant terms have been omitted.
As a result of the conditional independence assumptions made by factor graphs, note that each term in the summation can be minimized separately, and the Joint NLL baseline reduces to learning the parameters of each factor in isolation.

\subsubsection{Differentiable Filter Baselines}

As a baseline for end-to-end optimized probabilistic state estimation, we use differentiable recursive filters, as studied in \cite{haarnoja2016backprop, karkus2018particle,jonschkowski2018differentiable,kloss2018howtotrain,lee2020}.
Filters use the same inputs, transition models, and virtual sensor models as smoothers; we also compare both constant and heteroscedastic noise models.



\subsubsection{LSTM Baselines}
Following prior work on differentiable filtering~\cite{haarnoja2016backprop, karkus2018particle,jonschkowski2018differentiable,kloss2018howtotrain, lee2020}, we compare to long-short-term memory (LSTM)~\cite{HochSchm97} baselines.
We implement both unidirectional and bidirectional~\cite{schuster1997bidirectional} LSTM baselines, the latter of which, similar to a smoother framework, conditions each state estimate on both prior and subsequent observations.
LSTMs are passed both the input image frames and the ground-truth initial state of the system.

\subsection{Evaluation Metrics}
For evaluation on the visual tracking task, we follow \cite{haarnoja2016backprop,kloss2018howtotrain} and evaluate an RMSE trajectory recovery error in pixels.

For the visual odometry task, we follow \cite{haarnoja2016backprop,kloss2018howtotrain,jonschkowski2018differentiable,karkus2018particle} and only compare the final timestep against ground truth, using subsequences of length 100.
In this task, the estimate will drift over time.
For short subsequences, the final estimate is a good proxy to evaluate the estimation quality over the entire trajectory.
We report two terms: (i) {\em Translational (m/m)\/}, the positional error at the last time step normalized by ground-truth distance traveled, and (ii) {\em Rotational (deg/m)\/}, angular error at the last time step normalized by distance traveled.

\subsection{Loss Functions}

Smoothers are trained using either the joint NLL loss (Equation~\ref{eq:nll}) or the end-to-end surrogate loss (Equation~\ref{eq:mse_loss}).
The end-to-end trained visual tracking smoothers are trained via supervision on position only.
For the KITTI visual odometry smoothers, we report separate results for end-to-end supervision on position MSE (X and Y, weighted equally) and velocity MSE (linear and angular, normalized).

All filtering baselines are trained end-to-end, with supervision on the posterior distribution.
They match \cite{kloss2018howtotrain} and use either an end-to-end MSE (visual tracking, visual odometry) or end-to-end NLL (visual tracking) losses.
Both losses directly supervise posterior distributions from the filter.

LSTM baselines directly supervise position for the visual tracking task, and both position and orientation for the odometry task.

\section{Experimental Evaluation}
\label{sec:experimental_evaluation}

\begin{figure}[t!]
    \centering
    \includegraphics[width=\linewidth]{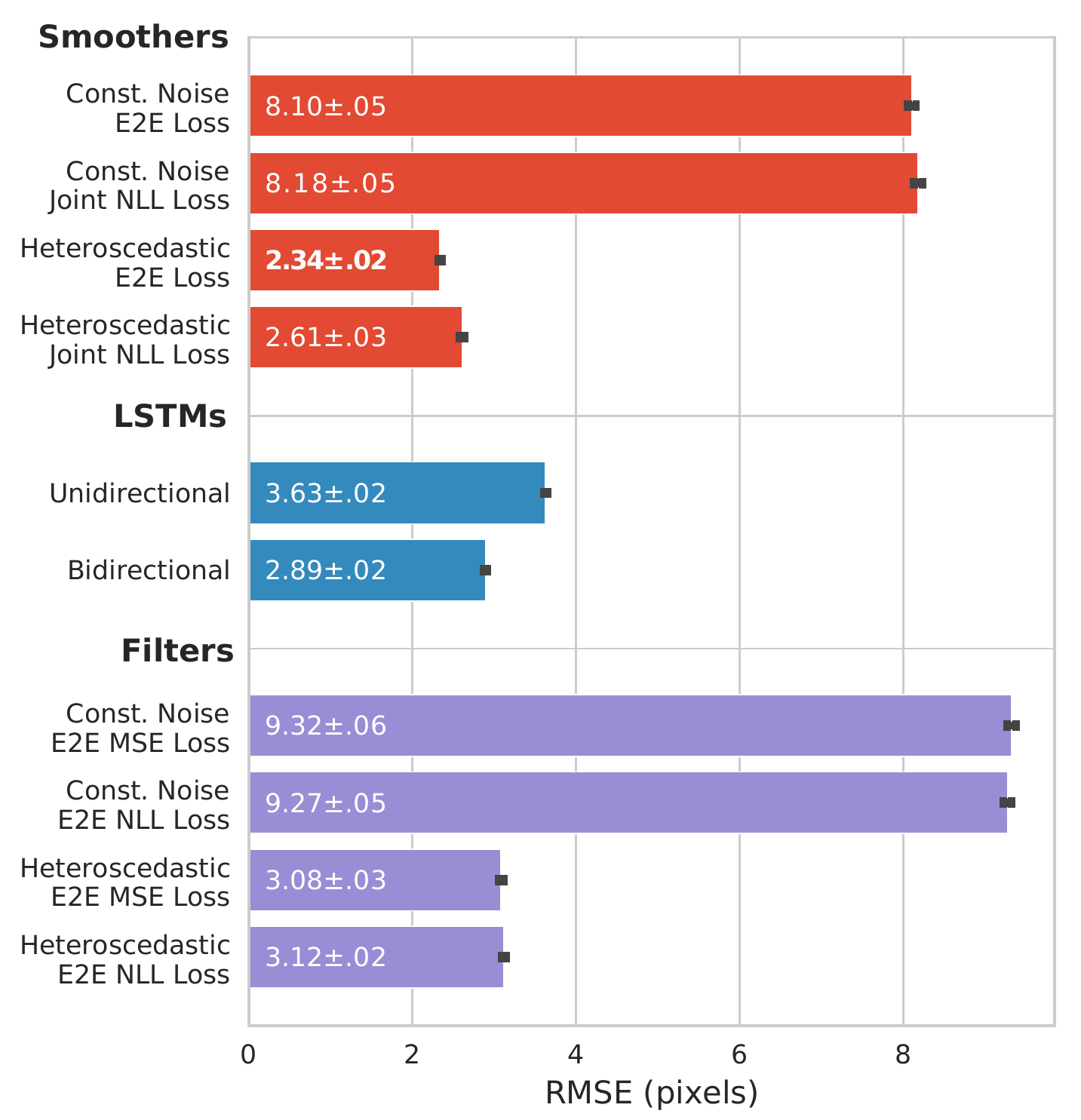}
    \caption{\label{fig:disk_bar}
        \textbf{Validation set errors on the visual tracking task.}
        We report mean and standard error metrics for smoothers, LSTMs, and Kalman filters across each of the 10 folds.
        A smoother trained using the proposed end-to-end surrogate loss outperforms all baselines.
    }
\end{figure}

\begin{table}[]
\scriptsize
\begin{tabular}{@{}llll@{}}
\toprule
\multicolumn{2}{c}{Model Type} & \multicolumn{1}{c}{m/m} & \multicolumn{1}{c}{deg/m} \\ \midrule
\multirow{3}{*}{\begin{tabular}[c]{@{}l@{}}Smoothers \\ Heteroscedastic\end{tabular}}
 & \multicolumn{1}{l|}{E2E Loss, Vel} & .1478 $\pm$ .0091 & \textbf{.0720} $\pm$ \textbf{.0059}\\
 & \multicolumn{1}{l|}{E2E Loss, Pos} & \textbf{.1457} $\pm$ \textbf{.0096} & .0762 $\pm$ .0067\\
 & \multicolumn{1}{l|}{Joint NLL Loss} & .1694 $\pm$ .0105 & .0724 $\pm$ .0057 \\
 \midrule
\multirow{3}{*}{\begin{tabular}[c]{@{}l@{}}Smoothers \\ Const. Noise\end{tabular}}
 & \multicolumn{1}{l|}{E2E Loss, Vel} & .1692 $\pm$ .0104 & .0724 $\pm$ .0057 \\
 & \multicolumn{1}{l|}{E2E Loss, Pos} & .1692 $\pm$ .0105 & .0724 $\pm$ .0057 \\
 & \multicolumn{1}{l|}{Joint NLL Loss} & .1695 $\pm$ .0105 & .0724 $\pm$ .0057 \\ 
 \midrule
\multirow{4}{*}{Baselines}
 & \multicolumn{1}{l|}{EKF, Heteroscedastic} & .1683 $\pm$ .0117 & .0743 $\pm$ .0066 \\
 & \multicolumn{1}{l|}{EKF, Const. Noise} & .1723 $\pm$ .0114 & .0745 $\pm$ .0065 \\
 & \multicolumn{1}{l|}{LSTM (unidirectional)} & .7575 $\pm$ .0199 & .3779 $\pm$ .0539 \\
 & \multicolumn{1}{l|}{LSTM (bidirectional)} & .7773 $\pm$ .0142 & .4773 $\pm$ .0743 \\
 \midrule
\multirow{3}{*}{\begin{tabular}[c]{@{}l@{}}Prior Work~\cite{kloss2018howtotrain}\end{tabular}}
 & \multicolumn{1}{l|}{EKF, Const. Noise} & .2112 $\pm$ .0342 & .0766 $\pm$ .0069 \\
 & \multicolumn{1}{l|}{UKF, Const. Noise} & .1800 $\pm$ .0233 & .0799 $\pm$ .0083 \\
 & \multicolumn{1}{l|}{LSTM (with dynamics)} & .5382 $\pm$ .0567 & .0788 $\pm$ .0083 \\
\bottomrule
\end{tabular}

\caption{\label{table:kitti}
    \textbf{Evaluation results on the KITTI visual odometry task.}
    As with the visual tracking task, end-to-end optimized smoothers with heteroscedastic noise model outperforms alternatives.
    Performance differences between our baselines and \citet{kloss2018howtotrain} can be attributed to implementation details such as data normalization and neural network weight initialization.
    The LSTM architecture from \cite{kloss2018howtotrain} also incorporates analytical dynamics information, which we exclude in our baselines.
}
\end{table}

\subsection{Overview}
By integrating the structure of a probabilistic smoother with end-to-end optimization, this work aims to provide performance benefits over (i) purely learned models such as LSTMs, which contain less problem-specific structure, (ii) state estimators with models that are trained outside of the overall estimator, which do not explicitly supervise end-to-end estimation error, and (iii) differentiable filtering approaches, which combine problem-specific structure and end-to-end optimization but suffer from limitations intrinsic to recursive filtering.
We study this hypothesis using the tasks and baselines detailed in Section~\ref{sec:experimental_setup}.
All experiments are cross-validated over 10 folds.

\subsection{Overall Estimation Accuracy}
Validation accuracies can be found in Figure~\ref{fig:disk_bar} for the visual tracking task and in Table~\ref{table:kitti} for the visual odometry task.
The proposed smoothing approaches outperform baselines on both tasks.
Consistent with prior work~\cite{haarnoja2016backprop,kloss2018howtotrain}, LSTMs result in the highest tracking errors of our evaluated models.

The results also highlight the importance of heteroscedastic noise models, which consistently improve tracking performance when compared to the constant noise models that are typically used when uncertainties are hand-tuned.
On the visual odometry task, a heteroscedastic noise model is also necessary to benefit from end-to-end training.
In the heteroscedastic noise model experiments, the Joint NLL Loss produces a tracking  error of .1694 m/m, which drops to .1453 m/m when the model is trained with an end-to-end position loss.
When a constant noise model is used, however, the Joint NLL and end-to-end losses produce nearly identical results: 0.1695 m/m and 0.1692 m/m respectively.

%
%
%
%

\subsection{Analyzing uncertainties}

\begin{figure}[t]
    \centering
    \vspace{0.1em}
    \includegraphics[width=3.5in]{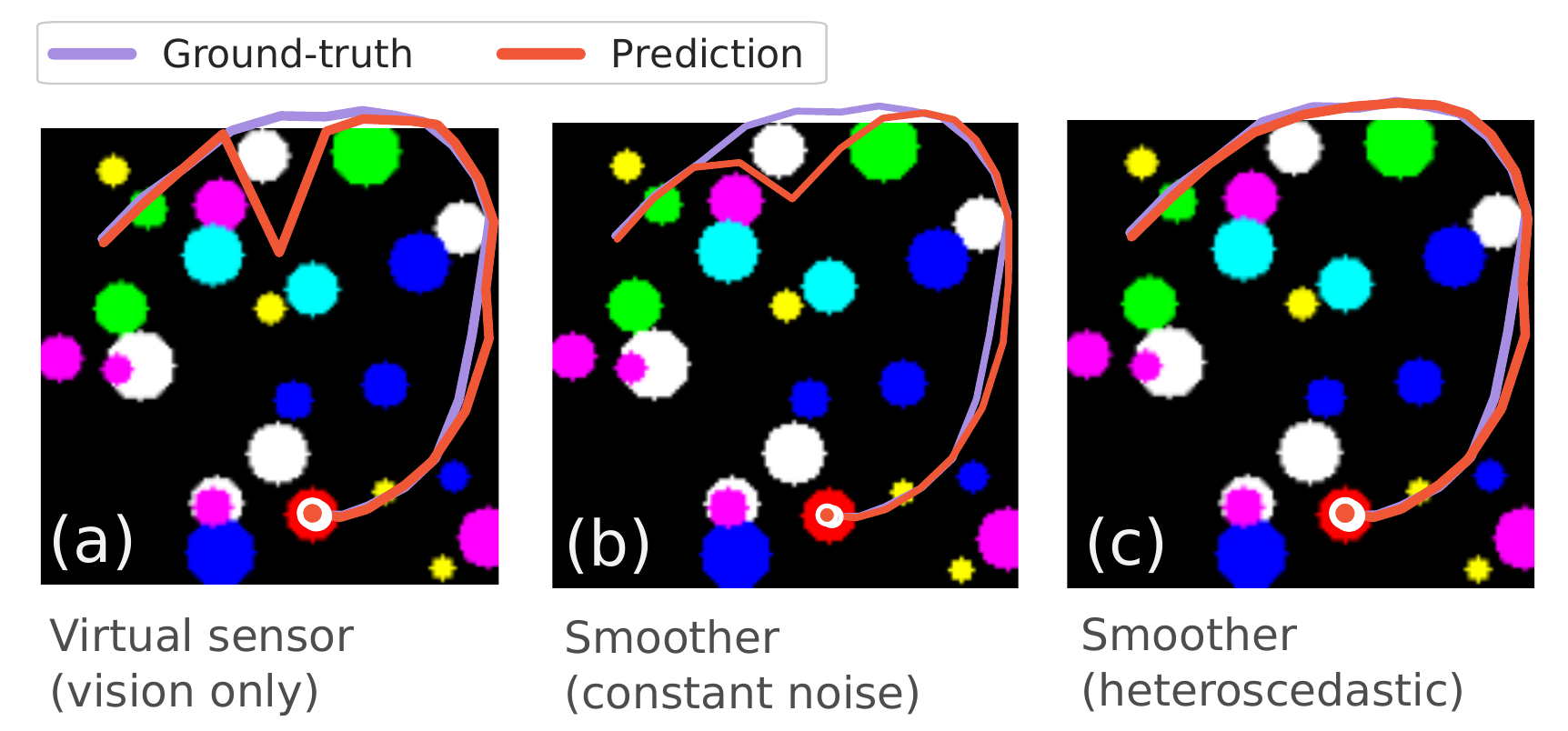}
    \vspace{0.1em}
    \caption{\label{fig:disk_visualization}
        \textbf{Visual tracking trajectories overlayed on a first-timestep observation.}
        We plot \textbf{(a)} predictions from the virtual sensor of a learned vision factor, \textbf{(b)} a smoothing output after learning constant virtual sensor noise models, and \textbf{(c)} a smoothing output after learning heteroscedastic noise models.
        A smoother with a heteroscedastic noise model is able to correctly reason about erroneous visual measurements when the disk exits the frame.
    }
\end{figure}

\begin{figure}[t]
    \centering
    \vspace{0.1em}
    \includegraphics[width=3.3in]{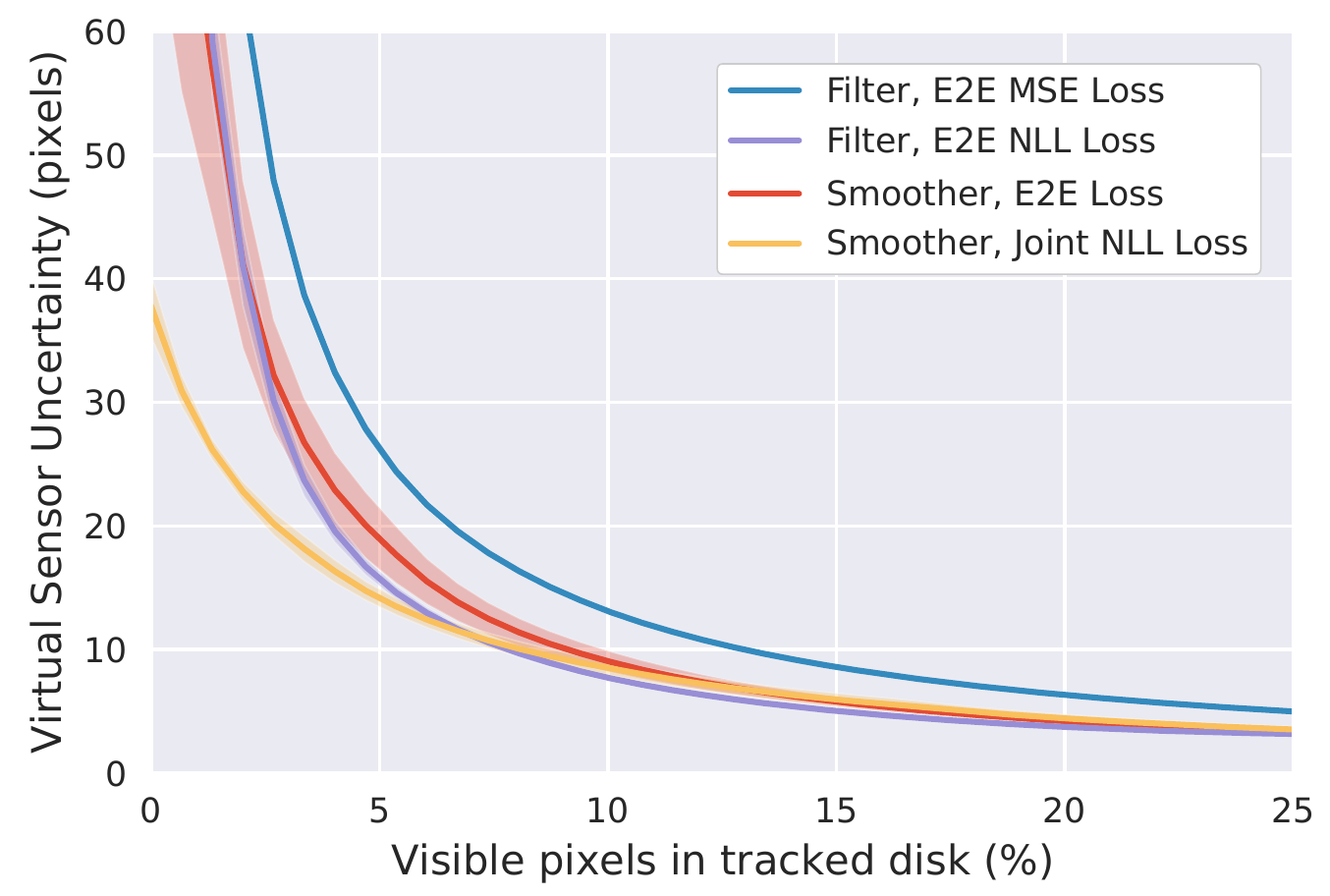}
    \vspace{0.1em}
    \caption{\label{fig:disk_uncertainty_plot}
        \textbf{Outputs of heteroscedastic noise models learned for the visual tracking task.}
        After training, the three end-to-end losses (Smoother E2E, EKF MSE, and EKF NLL) produce higher uncertainties than the Joint NLL baseline when the disk is occluded.
    }
\end{figure}
By leveraging the underlying probabilistic structure, end-to-end-optimized smoothers enable the direct inspection of uncertainties. 
Not only can this be useful for downstream decision-making, but uncertainties can also be used to interpret the underlying mechanics of learned state estimators.

Figure \ref{fig:disk_uncertainty_plot} visualizes learned heteroscedastic noise models for virtual sensors trained on the visual tracking task.
Compared to the Joint NLL Loss baseline, approaches that are trained end-to-end assign higher uncertainties to measurements when the disk is occluded.
This highlights the reason that end-to-end training is important for this task.
When the vision factor is probabilistically trained outside of the end-to-end context (ie using the Joint NLL Loss), the Gaussian noise model learns an uncertainty equivalent to the expected absolute error of virtual sensor predictions.
The uncertainty learned for vision factors, therefore, is bound by the dimensions of the area that the disk traverses.
This is optimal only when the errors are truly zero-mean Gaussians, which is rarely true in practice. 
In contrast, end-to-end methods are able to correctly reason that, when placed in context of the overall state estimator, vision factors provide no contribution when the tracked disk is fully occluded.
This allows the state estimator to completely ignore visual inputs, and instead rely solely on the dynamics model.

\subsection{Transferring noise models}
A key motivation for our approach is the ability to learn system models that are specifically optimized for the smoothing task that they are deployed in.
Filters and smoothers often make different sets of assumptions.
Therefore, a model that is optimized for filtering is not necessarily the same as a model that is optimized for smoothing.
To quantify this intuition, we evaluate the transfer of heteroscedastic noise profiles between state estimation contexts for the visual odometry task.
Noise models are learned using end-to-end MSE losses in filters and evaluated in both filters and smoothers; similarly, noise models are learned in smoothers and evaluated in both smoothers and filters.
Results are shown in Table \ref{table:noise_transfer}.

\begin{table}[]
\centering 
\begin{tabular}{@{}l|l|l@{}}
\toprule
& Tested on Filter & Tested on Smoother \\ \midrule
Trained on Filter & \textbf{.1683} $\pm$ \textbf{.0117}  & .1669 $\pm$ .0034\\
Trained on Smoother & .1705 $\pm$ .0045 & \textbf{.1457} $\pm$ \textbf{.0096}\\
\bottomrule
\end{tabular}
\caption{\label{table:noise_transfer}
    \textbf{Heteroscedastic noise model transfer results on the visual odometry task, positional tracking errors in m/m.}
    Each evaluation context performs best when run on models from a matching training context.
}
\end{table}

The immediate pattern present in these results is that smoothers always outperform filters.
Regardless of where a noise model is originally learned, the additional information that smoothers incorporate when compared to a filter results in consistently lower errors in recovered trajectories.

For a fixed evaluation context (filtering or smoothing), we also observe that the best results are achieved from a matching training context.
Compared to an end-to-end learned smoother, a significant drop in performance occurs when using a filter-trained model within a factor graph-based smoother.
A similar pattern emerges when implanting smoother-trained noise models into a filter.
These results underscore the simple insight that drives the utility of our method: in order to make the most out of available training data, it is important for model parameters to be learned in the same context that they are evaluated in.

\section{Conclusion}
In this work, we propose and demonstrate an approach for integrating optimization-based smoothing with end-to-end learning.
By backpropagating gradients through a fixed number of nonlinear optimization steps, learned smoothers can outperform both LSTMs and differentiable filters, while retaining the advantages of probabilistic state estimation.

There are many directions for extending these results.
Immediate applications exist where standard factor-graph based methods have outperformed filtering, which includes state estimation in SLAM, robot locomotion, manipulation, and tracking.
Opportunities are also presented by additional sensing modalities, such as audio or tactile feedback, where factor graphs have in part been constrained by the difficulty of analytical factor design.

\section{Acknowledgements}
We would like to thank Rika Antonova, Kevin Zakka, Angelina Wang, and Philipp Wu for insightful discussions and feedback.

This work is partially supported by the Toyota Research Institute (TRI) and Google. This article solely reflects the opinions and conclusions of its authors and not TRI, Google, or any entity associated with TRI or Google.

{
    \footnotesize
    \bibliographystyle{IEEEtranN}
    \bibliography{references}

\begin{thebibliography}{41}
\providecommand{\natexlab}[1]{#1}
\providecommand{\url}[1]{#1}
\csname url@samestyle\endcsname
\providecommand{\newblock}{\relax}
\providecommand{\bibinfo}[2]{#2}
\providecommand{\BIBentrySTDinterwordspacing}{\spaceskip=0pt\relax}
\providecommand{\BIBentryALTinterwordstretchfactor}{4}
\providecommand{\BIBentryALTinterwordspacing}{\spaceskip=\fontdimen2\font plus
\BIBentryALTinterwordstretchfactor\fontdimen3\font minus
  \fontdimen4\font\relax}
\providecommand{\BIBforeignlanguage}[2]{{%
\expandafter\ifx\csname l@#1\endcsname\relax
\typeout{** WARNING: IEEEtranN.bst: No hyphenation pattern has been}%
\typeout{** loaded for the language `#1'. Using the pattern for}%
\typeout{** the default language instead.}%
\else
\language=\csname l@#1\endcsname
\fi
#2}}
\providecommand{\BIBdecl}{\relax}
\BIBdecl

\bibitem[Thrun et~al.(2005)Thrun, Burgard, and Fox]{thrun2005probrob}
S.~Thrun, W.~Burgard, and D.~Fox, \emph{Probabilistic Robotics (Intelligent
  Robotics and Autonomous Agents)}.\hskip 1em plus 0.5em minus 0.4em\relax The
  MIT Press, 2005.

\bibitem[Dellaert and Kaess(2017)]{FactorGraphBook}
F.~Dellaert and M.~Kaess, \emph{Factor Graphs for Robot Perception}, 2017.

\bibitem[Li and Todorov(2007)]{RiskSensitive:Todorov:2007}
W.~Li and E.~Todorov, ``Iterative linearization methods for approximately
  optimal control and estimation of non-linear stochastic system,''
  \emph{International Journal of Control}, vol.~80, no.~9, pp. 1439--1453,
  2007.

\bibitem[Pont{\'o}n et~al.(2020)Pont{\'o}n, Schaal, and Righetti]{Ponton2020}
B.~Pont{\'o}n, S.~Schaal, and L.~Righetti, \emph{On the Effects of Measurement
  Uncertainty in Optimal Control of Contact Interactions}.\hskip 1em plus 0.5em
  minus 0.4em\relax Cham: Springer International Publishing, 2020, pp.
  784--799.

\bibitem[Eppner et~al.(2018)Eppner, Mart{\'\i}n-Mart{\'\i}n, and
  Brock]{eppner2018physics}
C.~Eppner, R.~Mart{\'\i}n-Mart{\'\i}n, and O.~Brock, ``Physics-based selection
  of informative actions for interactive perception,'' in \emph{2018 IEEE
  International Conference on Robotics and Automation (ICRA)}.\hskip 1em plus
  0.5em minus 0.4em\relax IEEE, 2018, pp. 7427--7432.

\bibitem[Kloss et~al.(2021)Kloss, Martius, and Bohg]{kloss2018howtotrain}
A.~Kloss, G.~Martius, and J.~Bohg, ``How to train your differentiable filter,''
  \emph{Autonomous Robot}, no.~45, pp. 561–--578, 2021.

\bibitem[Bradbury et~al.(2018)Bradbury, Frostig, Hawkins, Johnson, Leary,
  Maclaurin, Necula, Paszke, Vander{P}las, Wanderman-{M}ilne, and
  Zhang]{jax2018github}
\BIBentryALTinterwordspacing
J.~Bradbury, R.~Frostig, P.~Hawkins, M.~J. Johnson, C.~Leary, D.~Maclaurin,
  G.~Necula, A.~Paszke, J.~Vander{P}las, S.~Wanderman-{M}ilne, and Q.~Zhang,
  ``{JAX}: composable transformations of {P}ython+{N}um{P}y programs,'' 2018.
  [Online]. Available: \url{http://github.com/google/jax}
\BIBentrySTDinterwordspacing

\bibitem[Dellaert and Kaess(2006)]{dellaert2006squareroot}
F.~Dellaert and M.~Kaess, ``Square root sam: Simultaneous localization and
  mapping via square root information smoothing,'' \emph{The International
  Journal of Robotics Research}, vol.~25, no.~12, pp. 1181--1203, 2006.

\bibitem[Strasdat et~al.(2012)Strasdat, Montiel, and
  Davison]{strasdat2012visual}
H.~Strasdat, J.~M. Montiel, and A.~J. Davison, ``Visual slam: why filter?''
  \emph{Image and Vision Computing}, vol.~30, no.~2, pp. 65--77, 2012.

\bibitem[{Salas-Moreno} et~al.(2013){Salas-Moreno}, {Newcombe}, {Strasdat},
  {Kelly}, and {Davison}]{salasmoreno2013}
R.~F. {Salas-Moreno}, R.~A. {Newcombe}, H.~{Strasdat}, P.~H.~J. {Kelly}, and
  A.~J. {Davison}, ``Slam++: Simultaneous localisation and mapping at the level
  of objects,'' in \emph{2013 IEEE Conference on Computer Vision and Pattern
  Recognition}, 2013, pp. 1352--1359.

\bibitem[Leutenegger et~al.(2015)Leutenegger, Lynen, Bosse, Siegwart, and
  Furgale]{leutenegger2015}
\BIBentryALTinterwordspacing
S.~Leutenegger, S.~Lynen, M.~Bosse, R.~Siegwart, and P.~Furgale,
  ``Keyframe-based visual–inertial odometry using nonlinear optimization,''
  \emph{Int. J. Rob. Res.}, vol.~34, no.~3, p. 314–334, Mar. 2015. [Online].
  Available: \url{https://doi.org/10.1177/0278364914554813}
\BIBentrySTDinterwordspacing

\bibitem[Whelan et~al.(2015)Whelan, Kaess, Johannsson, Fallon, Leonard, and
  McDonald]{whelan2015}
T.~Whelan, M.~Kaess, H.~Johannsson, M.~Fallon, J.~J. Leonard, and J.~McDonald,
  ``Real-time large-scale dense rgb-d slam with volumetric fusion,'' \emph{The
  International Journal of Robotics Research}, vol.~34, no. 4-5, pp. 598--626,
  2015.

\bibitem[Kaess et~al.(2008)Kaess, Ranganathan, and Dellaert]{isam}
M.~Kaess, A.~Ranganathan, and F.~Dellaert, ``{iSAM}: Incremental smoothing and
  mapping,'' \emph{IEEE Trans. on Robotics (TRO)}, vol.~24, no.~6, pp.
  1365--1378, Dec. 2008.

\bibitem[Kaess et~al.(2012)Kaess, Johannsson, Roberts, Ila, Leonard, and
  Dellaert]{isam2}
M.~Kaess, H.~Johannsson, R.~Roberts, V.~Ila, J.~J. Leonard, and F.~Dellaert,
  ``isam2: Incremental smoothing and mapping using the bayes tree,'' \emph{The
  International Journal of Robotics Research}, vol.~31, no.~2, pp. 216--235,
  2012.

\bibitem[{Yu} and {Rodriguez}(2018)]{yu2018}
K.-T. {Yu} and A.~{Rodriguez}, ``Realtime state estimation with tactile and
  visual sensing for inserting a suction-held object,'' in \emph{2018 IEEE/RSJ
  International Conference on Intelligent Robots and Systems (IROS)}, 2018, pp.
  1628--1635.

\bibitem[Lambert et~al.(2019)Lambert, Mukadam, Sundaralingam, Ratliff, Boots,
  and Fox]{lambert2019visuotactile}
A.~S. Lambert, M.~Mukadam, B.~Sundaralingam, N.~D. Ratliff, B.~Boots, and
  D.~Fox, ``Joint inference of kinematic and force trajectories with
  visuo-tactile sensing,'' in \emph{International Conference on Robotics and
  Automation, {ICRA} 2019, Montreal, QC, Canada, May 20-24, 2019}.\hskip 1em
  plus 0.5em minus 0.4em\relax {IEEE}, 2019, pp. 3165--3171.

\bibitem[Sodhi et~al.(2021{\natexlab{a}})Sodhi, Kaess, Mukadam, and
  Anderson]{GelSightFAIR}
P.~Sodhi, M.~Kaess, M.~Mukadam, and S.~Anderson, ``Learning tactile models for
  factor graph-based state estimation,'' in \emph{2021 IEEE International
  Conference on Robotics and Automation}, 2021, submitted.

\bibitem[P{\"o}schmann et~al.(2020)P{\"o}schmann, Pfeifer, and
  Protzel]{Pschmann2020FactorGB}
J.~P{\"o}schmann, T.~Pfeifer, and P.~Protzel, ``Factor graph based 3d
  multi-object tracking in point clouds,'' \emph{2020 IEEE/RSJ International
  Conference on Intelligent Robots and Systems (IROS)}, pp. 10\,343--10\,350,
  2020.

\bibitem[Haarnoja et~al.(2016)Haarnoja, Ajay, Levine, and
  Abbeel]{haarnoja2016backprop}
T.~Haarnoja, A.~Ajay, S.~Levine, and P.~Abbeel, ``Backprop kf: Learning
  discriminative deterministic state estimators,'' in \emph{Advances in Neural
  Information Processing Systems}, 2016, pp. 4376--4384.

\bibitem[Karkus et~al.(2018)Karkus, Hsu, and Lee]{karkus2018particle}
P.~Karkus, D.~Hsu, and W.~S. Lee, ``Particle filter networks with application
  to visual localization,'' in \emph{Conference on Robot Learning}, 2018, pp.
  169--178.

\bibitem[Jonschkowski et~al.(2018)Jonschkowski, Rastogi, and
  Brock]{jonschkowski2018differentiable}
R.~Jonschkowski, D.~Rastogi, and O.~Brock, ``{Differentiable Particle Filters:
  End-to-End Learning with Algorithmic Priors},'' in \emph{{Proceedings of
  Robotics: Science and Systems (RSS)}}, 2018.

\bibitem[{Lee} et~al.(2020){Lee}, {Yi}, {Mart\'in-Mart\'in}, {Savarese}, and
  {Bohg}]{lee2020}
M.~A. {Lee}, B.~{Yi}, R.~{Mart\'in-Mart\'in}, S.~{Savarese}, and J.~{Bohg},
  ``Multimodal sensor fusion with differentiable filters,'' in \emph{2020
  IEEE/RSJ International Conference on Intelligent Robots and Systems (IROS)},
  2020, pp. 10\,444--10\,451.

\bibitem[Zachares et~al.(2021)Zachares, Lee, Lian, and
  Bohg]{zachares2021interpreting}
P.~A. Zachares, M.~A. Lee, W.~Lian, and J.~Bohg, ``Interpreting contact
  interactions to overcome failure in robot assembly tasks,'' in \emph{2021
  IEEE International Conference on Robotics and Automation (ICRA)}.\hskip 1em
  plus 0.5em minus 0.4em\relax Xi'an, China: IEEE, 2021.

\bibitem[Czarnowski et~al.(2020)Czarnowski, Laidlow, Clark, and
  Davison]{czarnowski2020deepfactors}
J.~Czarnowski, T.~Laidlow, R.~Clark, and A.~Davison, ``Deepfactors: Real-time
  probabilistic dense monocular slam,'' \emph{IEEE Robotics and Automation
  Letters}, vol.~5, pp. 721--728, 2020.

\bibitem[Rabiee and Biswas(2020)]{IV-SLAM}
S.~Rabiee and J.~Biswas, ``Iv-slam: Introspective vision for simultaneous
  localization and mapping,'' in \emph{Conference on Robot Learning}, 2020.

\bibitem[{Li} et~al.(2021){Li}, {Wu}, and
  {Kennedy}]{liwu2021replayovershooting}
A.~{Li}, P.~{Wu}, and M.~{Kennedy}, ``{Replay Overshooting}: Learning
  stochastic latent dynamics with the extended kalman filter,'' in \emph{2021
  International Conference on Robotics and Automation (ICRA)}, Xi'an, China,
  2021.

\bibitem[{Barratt} and {Boyd}(2020)]{barratt2020fitting}
S.~T. {Barratt} and S.~P. {Boyd}, ``Fitting a kalman smoother to data,'' in
  \emph{2020 American Control Conference (ACC)}, 2020, pp. 1526--1531.

\bibitem[Tang and Tan(2019)]{tang2018banet}
C.~Tang and P.~Tan, ``{BA}-net: Dense bundle adjustment networks,'' in
  \emph{International Conference on Learning Representations}, 2019.

\bibitem[{Jatavallabhula} et~al.(2020){Jatavallabhula}, {Iyer}, and
  {Paull}]{gradslam2020}
K.~M. {Jatavallabhula}, G.~{Iyer}, and L.~{Paull}, ``$\nabla$slam: Dense slam
  meets automatic differentiation,'' in \emph{2020 IEEE International
  Conference on Robotics and Automation (ICRA)}, 2020, pp. 2130--2137.

\bibitem[von Stumberg et~al.(2020)von Stumberg, Wenzel, Khan, and
  Cremers]{gn-net-2020}
L.~von Stumberg, P.~Wenzel, Q.~Khan, and D.~Cremers, ``{GN-Net}: The
  gauss-newton loss for multi-weather relocalization,'' \emph{{IEEE} Robotics
  and Automation Letters ({RA-L})}, vol.~5, no.~2, pp. 890--897, 2020.

\bibitem[Grefenstette et~al.(2019)Grefenstette, Amos, Yarats, Htut, Molchanov,
  Meier, Kiela, Cho, and Chintala]{grefenstette2019generalized}
E.~Grefenstette, B.~Amos, D.~Yarats, P.~M. Htut, A.~Molchanov, F.~Meier,
  D.~Kiela, K.~Cho, and S.~Chintala, ``Generalized inner loop meta-learning,''
  \emph{arXiv preprint arXiv:1910.01727}, 2019.

\bibitem[{Bhardwaj} et~al.(2020){Bhardwaj}, {Boots}, and
  {Mukadam}]{bhardwaj2020dgpmp}
M.~{Bhardwaj}, B.~{Boots}, and M.~{Mukadam}, ``Differentiable gaussian process
  motion planning,'' in \emph{2020 IEEE International Conference on Robotics
  and Automation (ICRA)}, 2020, pp. 10\,598--10\,604.

\bibitem[Karkus et~al.(2019)Karkus, Ma, Hsu, Kaelbling, Lee, and
  Lozano{-}P{\'{e}}rez]{KarkusMHKLL19}
P.~Karkus, X.~Ma, D.~Hsu, L.~P. Kaelbling, W.~S. Lee, and
  T.~Lozano{-}P{\'{e}}rez, ``Differentiable algorithm networks for composable
  robot learning,'' in \emph{Robotics: Science and Systems XV, University of
  Freiburg, Freiburg im Breisgau, Germany, June 22-26, 2019}, A.~Bicchi,
  H.~Kress{-}Gazit, and S.~Hutchinson, Eds., 2019.

\bibitem[{Kümmerle} et~al.(2011){Kümmerle}, {Grisetti}, {Strasdat},
  {Konolige}, and {Burgard}]{kummerle2011g2o}
R.~{Kümmerle}, G.~{Grisetti}, H.~{Strasdat}, K.~{Konolige}, and W.~{Burgard},
  ``G2o: A general framework for graph optimization,'' in \emph{2011 IEEE
  International Conference on Robotics and Automation}, 2011, pp. 3607--3613.

\bibitem[{Olson} et~al.(2006){Olson}, {Leonard}, and {Teller}]{OlsonGraph2006}
E.~{Olson}, J.~{Leonard}, and S.~{Teller}, ``Fast iterative alignment of pose
  graphs with poor initial estimates,'' in \emph{Proceedings 2006 IEEE
  International Conference on Robotics and Automation, 2006. ICRA 2006.}, 2006,
  pp. 2262--2269.

\bibitem[Sol\`a et~al.(2018)Sol\`a, Deray, and Atchuthan]{sola2018microlie}
J.~Sol\`a, J.~Deray, and D.~Atchuthan, ``A micro {L}ie theory for state
  estimation in robotics,'' \url{http://arxiv.org/abs/1812.01537}, {Institut de
  Rob\`otica i Inform\`atica Industrial}, Barcelona, Tech. Rep. IRI-TR-18-01,
  2018.

\bibitem[Hochreiter and Schmidhuber(1997)]{HochSchm97}
S.~Hochreiter and J.~Schmidhuber, ``Long short-term memory,'' \emph{Neural
  Computation}, vol.~9, no.~8, pp. 1735--1780, 1997.

\bibitem[Schuster and Paliwal(1997)]{schuster1997bidirectional}
M.~Schuster and K.~K. Paliwal, ``Bidirectional recurrent neural networks,''
  \emph{IEEE transactions on Signal Processing}, vol.~45, no.~11, pp.
  2673--2681, 1997.

\bibitem[Indelman et~al.(2012)Indelman, Williams, Kaess, and
  Dellaert]{indelman2012factor}
V.~Indelman, S.~Williams, M.~Kaess, and F.~Dellaert, ``Factor graph based
  incremental smoothing in inertial navigation systems,'' in \emph{2012 15th
  International Conference on Information Fusion}.\hskip 1em plus 0.5em minus
  0.4em\relax IEEE, 2012, pp. 2154--2161.

\bibitem[Jian et~al.(2012)Jian, Balcan, and Dellaert]{jian2012generalized}
Y.-D. Jian, D.~C. Balcan, and F.~Dellaert, ``Generalized subgraph
  preconditioners for large-scale bundle adjustment,'' in \emph{Outdoor and
  Large-Scale Real-World Scene Analysis}.\hskip 1em plus 0.5em minus
  0.4em\relax Springer, 2012, pp. 131--150.

\bibitem[Sodhi et~al.(2021{\natexlab{b}})Sodhi, Dexheimer, Mukadam, Anderson,
  and Kaess]{sodhi2021leo}
P.~Sodhi, E.~Dexheimer, M.~Mukadam, S.~Anderson, and M.~Kaess, ``Leo: Learning
  energy-based models in graph optimization,'' 2021.

\end{thebibliography}
}

\clearpage

\newpage
\appendix

\subsection{Advantages of Smoothing}
\label{appendix:whysmoothing}
While Kalman filters are optimal for simple linear Markov models, abandoning the measurement history in nonlinear systems leads to the accumulation of errors from approximations such as model linearization (in Extended Kalman filters) or sigma point belief propagation (in Unscented Kalman filters).
As a result, Kalman-type filters risk divergence in applications where these system approximations are highly state-dependent~\cite{yu2018}.
In contrast, a factor graph-based smoothing approach gives a user the flexibility to either maintain a full history of inputs or a partial sliding window to balance error reduction with computational efficiency \cite{indelman2012factor}. This results in improved robustness and reduced sensitivity to initialization.

Filtering-based approaches are highly efficient for problems with small state spaces.
However, they are limited by the need for an explicit parameterization of the uncertainty of the posterior.
For particle filters, distributions are represented using a discrete set of weighted samples.
As the state dimensionality increases linearly, the required number of samples increases exponentially.
For Kalman-type filters, distributions are represented with a covariance or precision matrix (sometimes in square-root form).
These matrices are dense and grow in size with the square of the state dimensionality, while the runtime of matrix operations grows cubically and quickly becomes intractable for systems with large or dynamic state spaces \cite{strasdat2012visual}. 
In contrast, optimization-based smoothing can be robustly applied to large-scale systems because computations are inherently sparse, allowing efficient optimization via either sparse matrix factorization \cite{dellaert2006squareroot} or matrix-free conjugate gradient methods \cite{jian2012generalized}, and a runtime that grows linearly with state dimensionality \cite{strasdat2012visual}. 

\subsection{Experimental Setup}

\subsubsection{Visual Tracking}
\label{appendix:experiment_visual_tracking}
The visual tracking environment enables full control over the process noise and observation complexity of the underlying system.
At each timestep $t$, the state consists of a 2D position $\diskpos_t$ and a 2D velocity $\diskvel_t$.
To generate a dataset for this task, we subject states to discrete-time linear-Gaussian dynamics:
\begin{equation}
    \label{eq:disk_dynamics}
    \begin{split}
        \diskpos_{t + 1} &= \diskpos_{t} + \diskvel_t + \diskq_{\diskpos, t}\\
        \diskvel_{t + 1} &= \diskvel_{t} - f_p \diskpos_{t} - f_d v_t^2 \text{sign}(v_t) + \diskq_{\diskvel, t}\\
    \end{split}
\end{equation}
where $\diskq$ is Gaussian process noise, $f_p$ is a linear spring coefficient, and $f_d$ is a drag coefficient.
For experiments, we use
$\diskq_{\diskvel, t} \sim \mathcal{N}(\mathbf{0}, \text{diag}(2, 2))$,
$\diskq_{\diskpos, t} \sim \mathcal{N}(\mathbf{0}, \text{diag}(0.1, 0.1))$,
$f_p = 0.05$,
and $f_d = 0.0075$.

As observations, state estimators have access to $120 \times 120 \times 3$ RGB images.
While the position of the red disc can be directly observed when it is visible, the state estimators are faced with frequent occlusions by 25 other distractor disks in the system, and the fact that the tracked red disk frequently leaves the visible frame.

We generate 5000 length-20 subsequences for this experiment, split into 10 folds for cross-validation.

\subsubsection{Visual Odometry}
\label{appendix:experiment_visual_odometry}
To generate data for this task, we reproduce the \texttt{kitti-10} dataset described in \cite{kloss2018howtotrain}.
The \texttt{kitti-10} dataset consists of 10 distinct trajectories of a car driving down different roads, which we can augment by (a) creating separate samples from each of two available camera feeds and (b) inserting both the original trajectory and a mirrored one into the dataset.
We perform 10-fold cross-validation by withholding 1 of these 10 trajectories for testing at a time. 

\subsection{Implementation Details}


\subsubsection{Nonlinear Optimizers}
We implement nonlinear optimization steps using Gauss-Newton updates with $K = 5$ for computing surrogate losses, and Levenberg-Marquardt for evaluating trained models.
Linear subproblems are solved with the Jacobi-preconditioned conjugate gradient method at train time, and a sparse Cholesky factorization during evaluation.
Note that optimizer convergence is trivial for the visual tracking task, as all factor residuals are linear.

\begin{table}[h!]
\centering 
\begin{tabular}{@{}l@{}}
\toprule
(1) 3x conv(3x3, 32, relu), conv(3x3, 32) \\
(2) Channel-wise max pool \\
(3) Concatenate width-mean pool and height-mean pool \\
(4) fc(32, relu) \\ 
(5) fc(2) \\
\bottomrule
fc: fully connected, conv: convolution
\end{tabular}
\caption{\textbf{Virtual sensor architecture for visual tracking position estimates.} Outputs are X/Y position estimates.}
\label{table:architecture_tracking}
\vspace{-3mm}
\end{table}

\subsubsection{Network Architectures}
\label{appendix:network_architectures}
The virtual sensor architecture for visual tracking can be found in Table~\ref{table:architecture_tracking}.
For heteroscedastic noise, we have a separate network that passes the \# of visible red pixels to 4x fc(64, relu) and fc(1), which then returns a square-root precision. 

The implementation of the LSTMs for the visual tracking task follows the same architecture as the virtual sensors, but expands the hidden unit count of the final dense layer to 32.
The output is passed to an LSTM cell with hidden state size 64, followed by a set of dense decoder layers. 
The bidirectional LSTM has the same architecture as the LSTM, but introduces a second set of LSTM cells.

For the KITTI visual odometry task, we follow the same virtual sensor network architecture as~\cite{kloss2018howtotrain}.
For both factor graphs and filters, we split our training set into two parts: the first for pretraining, which is used for learning velocity predictions via a simple MSE loss, and the second for end-to-end training, which is used to learn uncertainties.

\subsubsection{Initialization \& Pretraining}
Consistent with pretraining procedures used for differentiable filters \cite{kloss2018howtotrain}, we observed that initialization was crucial for end-to-end trained smoothers to converge correctly.
Virtual sensors are pretrained with an MSE loss on their mean outputs, and learned noise models for virtual sensors are biased using the pretrained models' RMSE metrics.

End-to-end training without pretraining or intentional initialization of networks parameters was prone to overfitting, local minima, and training stability issues; concurrent work has shown that these issues may be alleviated with energy-based training approaches~\citep{sodhi2021leo}.


\end{document}